\documentclass[journal]{IEEEtran}
\usepackage{epsfig}
\usepackage{graphicx}
\usepackage{comment}
\usepackage{amsmath,amssymb} % define this before the line numbering.
\usepackage{amsfonts}
\usepackage{array}
\usepackage{algorithm}
\usepackage{algorithmic}
\usepackage{cite}
\usepackage{booktabs}
\usepackage{float}
\usepackage{multirow}
\usepackage{color}
\usepackage{bm}
\usepackage{soul}
\soulregister\ref7 % 针对\ref命令
\soulregister\cite7 % 针对\cite命令

\newcommand\norm[1]{\left\lVert#1\right\rVert}

\begin{document}
\title{Spatio-Temporal Graph Dual-Attention Network for Multi-Agent Prediction and Tracking}

%\title{STG-DAT: Interaction-aware Trajectory Prediction via Wasserstein Graph dual-attention Network}

\author{Jiachen~Li, %~\IEEEmembership{Student Member,~IEEE,}
        Hengbo~Ma, %\IEEEmembership{Student Member,~IEEE,}
        Zhihao~Zhang, %~\IEEEmembership{Student Member,~IEEE,}
        Jinning~Li, %~\IEEEmembership{Student Member,~IEEE,}
        and~Masayoshi~Tomizuka,~\IEEEmembership{Life~Fellow,~IEEE}% <-this % stops a space
\thanks{J. Li, H. Ma, Z. Zhang, J. Li and M. Tomizuka are with the Department of Mechanical Engineering, University of California, Berkeley, CA 94720, USA (Email: \{\tt\small jiachen\_li, hengbo\_ma, zhihaozhang, jinning\_li, tomizuka\}@berkeley.edu)}% <-this % stops a space
}


\markboth{Journal of \LaTeX\ Class Files,~Vol.~X, No.~X, May~2020}%
{Shell \MakeLowercase{\textit{et al.}}: Bare Demo of IEEEtran.cls for IEEE Journals}



\maketitle

\begin{abstract}
An effective understanding of the environment and accurate trajectory prediction of surrounding dynamic obstacles are indispensable for intelligent mobile systems (e.g. autonomous vehicles and social robots) to achieve safe and high-quality planning when they navigate in highly interactive and crowded scenarios.
Due to the existence of frequent interactions and uncertainty in the scene evolution, it is desired for the prediction system to enable relational reasoning on different entities and provide a distribution of future trajectories for each agent.
In this paper, we propose a generic generative neural system (called STG-DAT) for multi-agent trajectory prediction involving heterogeneous agents. The system takes a step forward to explicit interaction modeling by incorporating relational inductive biases with a dynamic graph representation and leverages both trajectory and scene context information.
We also employ an efficient kinematic constraint layer applied to vehicle trajectory prediction. The constraint not only ensures physical feasibility but also enhances model performance.
Moreover, the proposed prediction model can be easily adopted by multi-target tracking frameworks. The tracking accuracy proves to be improved by empirical results.
The proposed system is evaluated on three public benchmark datasets for trajectory prediction, where the agents cover pedestrians, cyclists and on-road vehicles. The experimental results demonstrate that our model achieves better performance than various baseline approaches in terms of prediction and tracking accuracy.
\end{abstract}


% \fi
\IEEEpeerreviewmaketitle

\section{Introduction}
\begin{figure*}[!tbp]
    \centering
    \includegraphics[width=0.95\textwidth]{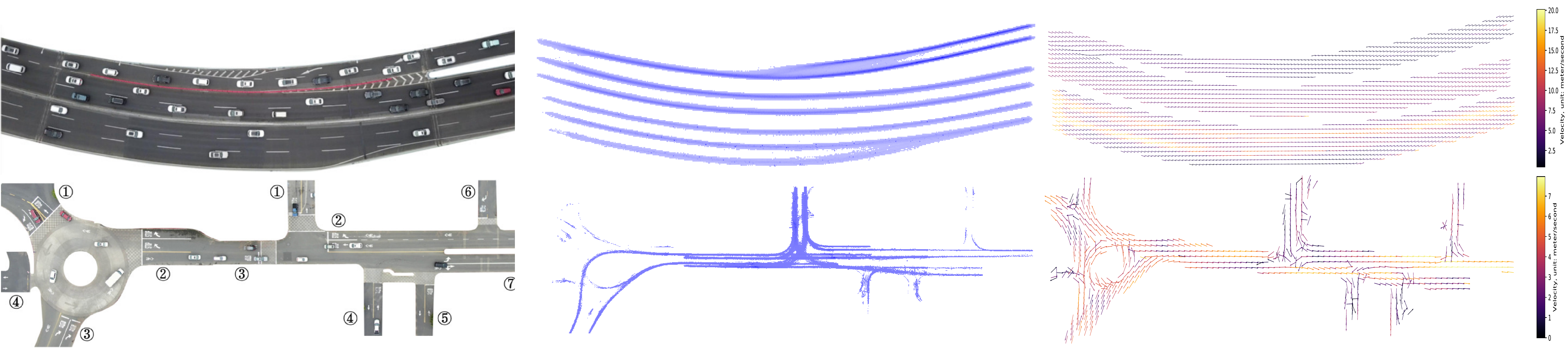}
    \caption{Typical traffic scenarios with large uncertainty and interactions among multiple entities. The left column is adopted from \cite{interactiondataset}. The upper figure in the first column was captured in a highway ramp merging scenario, where lane change behavior with negotiation happens frequently. The lower figure was captured in a roundabout and an unsignalized intersection scenario, where yielding and stopping behaviors happen frequently. The other two columns shows the occupancy density maps and the velocity fields of the scenarios, which are generated based on the training data to provide statistical context information.}
    \label{fig:scenarios}
\end{figure*}

In order to navigate safely in dense traffic scenarios or crowded areas full of vehicles and pedestrians, it is crucial for autonomous vehicles or mobile robots to forecast and track future behaviors of surrounding interactive agents accurately and efficiently \cite{lefevre2014survey}. For short-term prediction, it may be acceptable to use pure physics based methods. However, due to the uncertain nature of future situations, the system for long-term prediction is desired to not only allow for interaction modeling between different agents, but also to figure out traversable regions delimited by road layouts as well as right of way compliant to traffic rules.
Fig. \ref{fig:scenarios} illustrates several traffic scenarios where interaction happens frequently and the drivable areas are heavily defined by road geometries. For instance, at the entrance of roundabouts or unsignalized intersections, the future behavior of an entering vehicle highly depends on whether the conflicting vehicles would yield and leave enough space for it to merge. In addition, for vehicle trajectory prediction, kinematic constraints should be satisfied to make the trajectories feasible and smooth. 

There have been extensive studies on the prediction of a single target entity, which consider the influences of its surrounding entities \cite{hong2019rules,chai2019multipath,alahi2016social,gupta2018social}. However, such approaches only care about one-way interactions, ignoring the potential interactions in the opposite way. Recent works have tried to address this issue by simultaneously forecasting for multiple agents \cite{deo2018convolutional,deo2018would,zhao2019multi}. However, most of these methods employ concatenation or social pooling operations to blend the features of different agents without explicit relational reasoning. 
Moreover, they are not able to model higher-order interactions (indirect influences) beyond adjacent entities.
In this work, we take a step forward to model the interactions explicitly with a spatio-temporal graph representation and attention-based message passing rules. Our model enables permutation invariance and mutual effects between pairs of entities.

In the literature of deep learning methods, the soft attention mechanism is widely adopted in modeling spatial relations due to its flexibility and interpretability. The temporal relations are usually modeled by recurrent neural networks, which naturally attenuate the effects of distant history information.
One of the most related work is Social Attention \cite{vemula2018social}.
However, in some cases the earlier information may be also important.
Therefore, in order to figure out which of the other agents have the most significant influence on a certain agent, as well as the relative importance of different time steps, we propose a spatio-temporal graph attention mechanism which is applied to both topological and temporal features.

Besides the trajectory prediction task, the proposed approach can also be incorporated into multi-target tracking frameworks, due to its property of permutation invariance and flexibility on agent numbers. 
More specifically, it can serve as the process (prediction) model in the prior update of recursive Bayesian state estimation. The model has advantages in handling occlusion issues and missing observations, due to its capability of long-term prediction.
In this paper, we adopt the multi-target tracking framework proposed in \cite{li2019generic} and compare the performance of tracking with our model and other widely used models.

This paper is a significant extension of our prior work \cite{ma2019} where we presented a modified Wasserstein generative modeling method. This is adopted as the basis of training the proposed model in this paper. The method in \cite{ma2019} was only able to predict interactive behaviors of two vehicles in a single scenario, while the proposed approach is able to handle multiple, heterogeneous agents in different scenarios simultaneously.
The main contributions of this paper are summarized as follows:
\begin{itemize}
    \item We propose a multi-agent, generative trajectory forecasting system with relational reasoning on heterogeneous, interactive agents. The system is applied to predict pedestrian and vehicle trajectories across different scenarios.
    \item We propose a spatio-temporal dual-attention mechanism for representation learning on spatio-temporal dynamic graphs, which can figure out relative significance of the information about different surrounding agents at each time step.
    \item We incorporate an efficient kinematic constraint module similar to \cite{ma2019} to ensure physical feasibility for vehicle trajectory prediction. This constraint layer can not only smooth the trajectories and reduce prediction error, but also enhance the model robustness to noisy data.
    \item We validate the proposed system on multiple trajectory forecasting benchmark datasets. The approach achieves state-of-the-art prediction accuracy. The model also proves to enhance the multi-target tracking performance.
\end{itemize}

The remainder of the paper is organized as follows.
Section II provides a brief overview on state-of-the-art related research.
Section III introduces preliminary background of the proposed approach. 
Section IV presents a generic problem formulation for the trajectory prediction task. 
Section V illustrates the proposed forecasting system. 
In Section VI, the proposed system is applied to interactive trajectory prediction of pedestrians and vehicles using real-world benchmark datasets. The performance is compared with various baseline methods in terms of widely-used evaluation metrics. 
A comprehensive ablative analysis is also provided to illustrate the necessity of each component. The prediction model is also applied to a vehicle tracking framework.
Finally, Section VII concludes the paper.
The details of data preprocessing are introduced in the appendix.

\section{Related Work}
In this section, we provide a concise literature review on related research and illustrate the distinctions and advantages of the proposed generative trajectory prediction approach, which can also be leveraged by multi-object tracking frameworks to enhance tracking performance.

\subsection{Interaction-Aware Trajectory Prediction}

Extensive research has been conducted on trajectory prediction of humans and autonomous agents (e.g. on-road vehicles, mobile robots, etc). Early literature mainly introduced physics-based or rule-based approaches, such as state estimation techniques based on kinematic models. These methods do not consider mutual influence between intelligent agents and they are only able to perform well in short-term prediction tasks with limited model flexibility \cite{liu2016vehicle,scharcanski2010particle}. 
As machine learning techniques are studied more extensively, people began to employ learning-based models for prediction purpose, such as hidden Markov models \cite{wang2018learning}, Gaussian mixture models \cite{Jiachen_ITSC18-2}, dynamic Bayesian network \cite{kasper2012object}, and inverse reinforcement learning \cite{sun2018probabilistic}. 
In recent years, researchers have proposed various learning-based prediction models, which enables more flexibility and capacity to capture underlying interactive behavior patterns \cite{fernando2018soft+,lee2017desire,rudenko2019human,xu2018encoding,liang2019peeking,ma2019trafficpredict,li2019conditional,li2018development,huang2019uncertainty,su2019potential,ivanovic2019trajectron,chai2019multipath,rhinehart2019precog,li2020evolvegraph,shitao,sadeghian2019sophie,zhang2019sr,choi2020shared,park2020diverse,rhinehart2018r2p2}. The prediction hypotheses can be directly obtained from the model outputs.
However, with an end-to-end fashion, physical feasibility constraints are usually ignored in these methods, which may result in implausible forecast.
In this paper, we address multi-agent interaction modeling and introduce a probabilistic prediction system based on a deep generative framework. Our approach also explicitly considers the feasibility constraints of vehicles.

\subsection{Multi-Target Tracking and State Estimation}
Although many research have been conducted on end-to-end tracking through real-time detection in the computer vision community, here we only focus on the approaches based on recursive Bayesian state estimation, which consists of prior (prediction) update and posterior (measurement) update. 
One of the challenges in the tracking problem is noisy or missing observations (occlusions). Kinematics based models tend to be sensitive to measurement noise and lose tracking when the targets are occluded for a period. Moreover, they are not able to consider interactive behaviors between tracking targets since the target states are propagated independently. Some approaches to handle these issues were proposed in \cite{li2019generic}. 
In their work, a learning-based behavior model is employed to improve the tracking accuracy. However, the behavior model requires specific assumptions on agent numbers, agent roles and scenarios, which significantly restricts the applicability.
The prediction model proposed in this paper can serve as a strong alternative to the original behavior model, which brings more versatility and larger model capacity.

\subsection{Relational Reasoning and Graph Networks}

In general, the objective of relational reasoning is to reason about different entities and their relations from observed structured or unstructured data, such as image pixels \cite{wang2018non}, words or sentences \cite{lin2017structured}, human skeletons \cite{kipf2018neural} and interactive navigating agents \cite{zambaldi2018relational,choi2019looking,ma2020reinforcement}.
Popular techniques for relational reasoning and interaction modeling in earlier literature include, but are not limited to, social pooling mechanism \cite{alahi2016social}, convolutional pooling mechanism \cite{deo2018convolutional}, soft attention mechanism \cite{vemula2018social}, etc.
Recently, graph networks have proved to be effective for relational reasoning on graph-structured data, where there is no restriction on the message passing rules.
In traffic scenarios, a typical representation of the whole scene is to formulate a graph, where nodes are agents and edges are their relationships. Most existing works focused on the approximation function parameterized by deep neural network due to its high flexibility, which leads to graph neural networks (GNN). In this paper, we present a graph neural network with both topological and temporal attention mechanisms to capture underlying interaction patterns and jointly predict future behaviors.

\subsection{Deep Generative Models}
Our approach is also related to deep generative models, which have been widely applied to representation learning and distribution approximation tasks \cite{goodfellow2014generative, kingma2013auto}. 
One of the advantages of generative modeling lies in the data distribution learning without supervision. Coupled with highly flexible deep networks, deep generative models have achieved satisfying performance in image generation, style transfer, sequence synthesis tasks, etc. 
Despite the variational auto-encoder, a highly flexible latent variable model with encoder-decoder architecture, tries its best to make the posterior of the latent variable and its prior (usually a normal distribution) as similar as possible, the two distributions do not match well in many tasks. Hence, it breaks the consistency of the model. 
Also, although generative adversarial networks have achieved satisfying performance on image generation tasks, it usually suffers from mode collapse problems, especially when applied to sequential data under the conditional setting. 
In order to mitigate these drawbacks, The Wasserstein auto-encoder (WAE) \cite{tolstikhin2017wasserstein}, proposed from the optimal transport point of view and combined with information theory, encourages the consistency between the encoded latent distribution and the prior distribution. A variant of variational auto-encoder was proposed in \cite{zhao2017infovae} and a modified approach was proposed in our previous work \cite{ma2019}.
In this paper, we adopt the similar Wasserstein generative method in \cite{ma2019} as the basis of model training, and significantly extends pair-wise prediction to multi-agent prediction.

\section{Preliminaries}
In this section, we first provide a high-level summary of basics of graph neural networks.
Then, we concisely introduce the Wasserstein generative modeling proposed in \cite{ma2019}.

\subsection{Graph Neural Network}
The graph neural network is a type of deep learning models which are directly applied to graph structures. It naturally incorporates relational inductive bias into the model design.
In the context of graph neural network, most graphs are attributed (with node attributes and/or edge attributes and/or global attributes).
Generally, there are three basic operations in graph representation learning with GNN: edge update, node update and global update \cite{battaglia2018relational}. Note that the global update is optional, which is applied only if the graphs have global attributes.
More formally, denote the graph with $n$ nodes as $\mathcal{G}=\{\mathcal{V},\mathcal{E}\}$, where $\mathcal{V}=\{v_i, i \in \{1,...,n\}\}$ is a set of node attributes and $\mathcal{E}=\{e_{ij}, i,j \in \{1,...,n\}\}$ is a set of edge attributes. Denote $u$ as the global attribute. Then, the three update operation can be written as 
\begin{equation}
    \begin{aligned}
    e'_{ij} &= \phi^e(e_{ij}, v_i, v_j, u), \ \qquad \bar{e}'_i = \rho^{e\rightarrow v}(E'_i), \\
    v'_i &= \phi^v(\bar{e}'_i, v_i, u), \qquad\qquad \bar{e}'=\rho^{e \rightarrow u}(E'), \\
    u' &= \phi^u(\bar{e}', \bar{v}', u), \qquad\qquad \bar{v}'=\rho^{v\rightarrow u}(V'), 
    \end{aligned}
\end{equation}
where $E'_i=\{e'_{ij}, j\in N(i)\}$, $E'=\bigcup E'_i$, $V'=\{v'_i,i=1,...,n\}$, and $N(i)$ is the neighbors of node $i$.
$\phi^e(\cdot)$,$\phi^v(\cdot)$ and $\phi^u(\cdot)$ are neural networks.
$\rho^{e\rightarrow v}(\cdot)$, $\rho^{e \rightarrow u}(\cdot)$ and $\rho^{v\rightarrow u}(\cdot)$ are aggregation functions with the property of permutation invariance.

In this work, it is natural to represent intelligent agents as nodes, and their relation as edges.
We only apply edge update and node update, since there is no global attribute in our setup.

\subsection{Wasserstein Generative Modeling}
The Wasserstein distance is defined in a metric space $(\chi, \rho)$:
\begin{equation}
\label{eq:Wdistance}
\begin{aligned}
W_{\rho}(Q, P)=\sup_{||f||_\text{Lip}\le 1} \int{f(dQ-dP)},
\end{aligned}
\end{equation}
where $||f||_\text{Lip}$ is the Lipschitz constant of the function $f$. 
By the Kantorovich-Rubinstein duality, we can formulate Eq. (\ref{eq:Wdistance}) as an optimal transport problem, which is given by 
\begin{equation}
\label{eq:dual_Wdistance}
\begin{aligned}
W_{\rho}(Q, P) = \inf_{\mathbf{M}}\int\rho(x, x')d\mathbf{M}=\inf_{\mathbf{M}}\mathbf{E}_{M}[\rho],
\end{aligned}
\end{equation}
where $\mathbf{M}(x, x')$ is the coupling distribution of $x \sim P$, $x' \sim Q$, which is a probability measure for $\chi \times \chi$.

Define 
\begin{equation}
    p_{G}(x):=\int_{z}p_{G}(x|z)p_{z}(z)\text{d}z, \forall x \in \chi.
\end{equation}
Following the WAE \cite{tolstikhin2017wasserstein} with the assumption that $p_{G}(x|z)$ is deterministic, we have
\begin{equation}
\label{eq:WAE}
\begin{aligned}
\inf_{M\in\mathcal{P}(P_X, P_G)}\mathbf{E}_{M}[\rho(X, Y)]
=\inf_{Q_z=P_z}\mathbf{E}_{p}\mathbf{E}_{q(z|x)}[\rho(X, G(Z))],
\end{aligned}
\end{equation}
where  $X \sim P_X$, $Z \sim Q(Z|X)$ and $Q_z$ is the marginal distribution of $Z$. 
%Eq. (\ref{eq:WAE}) implies that we can introduce an auxiliary variable to factorize the joint distribution, so that we can find a measure of joint distribution which can minimize the expectation of  cost between two different random variables. 
%Since it is convenient to sample from the distribution $Q(Z|X)$ rather than $Q(z)$, we can modify the optimization problem to be 
% \begin{align}
% &\min_{Q(Z|X)\in \mathcal{Q}}\mathbf{E}_{P_{X}}\mathbf{E}_{Q(Z|X)}[\rho(X, G(Z))]\\
% &s.t. \: P_Z = \mathbf{E}_{P_X}[Q(Z|X)].
% \end{align}
% There is an upper bound of the KL divergence between $Q(z)$ and $P(z)$. 
% Therefore, by the Fubini's theorem, we have
% \begin{equation}
% \label{eq:KL}
% \begin{aligned}
% \mathcal{D}_{KL}(Q(Z)||P(Z)) \le \mathbf{E}_{p(x)}[\mathcal{D}_{KL}(Q(Z|X)||P(Z))].
% \end{aligned}
% \end{equation}
% The gap between LHS and RHS of (\ref{eq:KL}) is the mutual information between $X$ and $Z$. 
% Since we know the upper bound of the divergence between $Q$ and $P$, 
We relax the optimization problem as:
\begin{equation}
\begin{aligned}
&\min_{Q(Z|X)\in \mathcal{Q}}\mathbf{E}_{P_{X}}\mathbf{E}_{Q(Z|X)}[\rho(X, G(Z))]\\
s.t.&
\left\{
\begin{aligned}
&\mathbf{E}_{p(x)}[\mathcal{D}_\text{KL}(Q(Z|X)||P(Z))] \le \epsilon_1,\\
&\mathcal{D}(Q_Z, P_Z) \le \epsilon_2,
\end{aligned}
\right.
\end{aligned}
\end{equation}
where $\epsilon_1$ and $\epsilon_2$ are pre-defined constants.
Then, the dual optimal problem is formulated as
\begin{equation}
\label{eq:max_min}
\begin{aligned}
&\max_{\alpha, \beta}\min_{Q(Z|X)\in \mathcal{Q}}\mathbf{E}_{P_{X}}\mathbf{E}_{Q(Z|X)}[\rho(X, G(Z))]\\
+ &\alpha \mathbf{E}_{p(x)}[\mathcal{D}_\text{KL}(Q(Z|X)||P(Z))] + \beta \mathcal{D}(Q_Z, P_Z),
\end{aligned}
\end{equation}
where $\alpha$ and $\beta$ are chosen from a proper range.

After some algebra derivations, we obtain the following equivalent optimization problem
\begin{equation}
\begin{aligned}
&\left \{
\begin{aligned}
&\max I_{\phi}(x, z)\\
s.t.&
\left \{
\begin{aligned}
&\mathcal{D}_\text{KL}[p_{\phi}(z)||p(z)]\le \epsilon_1, \\
&\mathcal{D}_\text{KL}[p_{\phi}(x, z)||p_{\theta}(x, z)] \le \epsilon_2.\\
\end{aligned}
\right.
\end{aligned}
\right.\\
\Leftrightarrow &
\left \{
\begin{aligned}
\min_{0<1-\alpha<\beta} &-(1-\alpha) I_{\phi}(x, z) \\ +& (\beta+ \alpha-1)\mathcal{D}_\text{KL}[p_{\phi}(z)||p(z)]\\ + &\mathcal{D}_\text{KL}[p_{\phi}(x, z)||p_{\theta}(x, z)],\\
\end{aligned}
\right .
\end{aligned}
\end{equation}
where $I_{\phi}(x,z)$ is the mutual information between $x$ and $z$.

Please refer to our prior work \cite{ma2019} for more details on the derivation. 
The goal of both WAE and vanilla variational auto-encoder (VAE) is to learn the data distribution. According to the numerical experiment results in \cite{ma2019}, the vanilla VAE tends to learn a distribution with smaller variance and have mode collapse issue; while WAE is better at capturing the true data distribution. 
Also, WAE is able to learn a better latent representation than VAE due to the regularization terms.

\section{Problem Formulation}
The objective is to predict future trajectories for multiple interactive agents, based on their historical states and context information. The prediction system can be also incorporated into any multi-target tracking frameworks.
Without loss of generality, we assume $N$ agents are navigating in the observation area, which are divided into $M$ categories. In this work, the involved agents include vehicles, pedestrians and cyclists.
We denote a set of agent trajectories as
\begin{equation}
\begin{aligned}
\mathbf{T}^{1:T}=\{\bm{\tau}_i^{1:T}|\bm{\tau}_i^{k}=(x_i^k, y_i^k,v_i^k,\psi_i^k), \\
T=T_h+T_f,i=1,...,N\},
\end{aligned}
\end{equation}
where $T_h$ is the history horizon and $T_f$ is the forecasting horizon.
$(x_i^k,y_i^k)$ is the position, $v_i^k$ is the velocity, and $\psi_i^k$ is the heading angle of agent $i$ at time $k$. 
The coordinates can be either in the world space or image pixel space. 
We also denote a sequence of context information (raw images, semantic maps or the tensors which includes other relevant information) as
\begin{equation}
    \begin{aligned}
    \mathbf{C}_g^{1:T}&=\{\mathbf{c}^{1:T},T=T_h+T_f\} \ \text{(global)},\\
    \mathbf{C}_l^{1:T}&=\{\mathbf{c}_i^{1:T}, T=T_h+T_f, i=1,...,N\} \ \text{(local)},\\
    \end{aligned}
\end{equation}
which indicates components in the high-definition maps (e.g. road geometries, road lanes, drivable areas, traffic signs, etc).
The future information is accessible during training.
We aim to approximate the conditional distribution $p(\mathbf{T}^{T_h+1:T_h+T_f} | \mathbf{T}^{1:T_h}, \mathbf{C}^{1:T_h})$.
The number of involved agents can be flexible in different cases.
In the multi-target tracking tasks, the prediction model is iteratively applied.

\section{Method: STG-DAT}
In this section, we first provide an overview of the key modules and the architecture of the proposed generative trajectory prediction system. The detailed model design of each module will then be further illustrated.

\subsection{System Overview}
\begin{figure*}[!tbp]
    \centering
    \includegraphics[width=\textwidth]{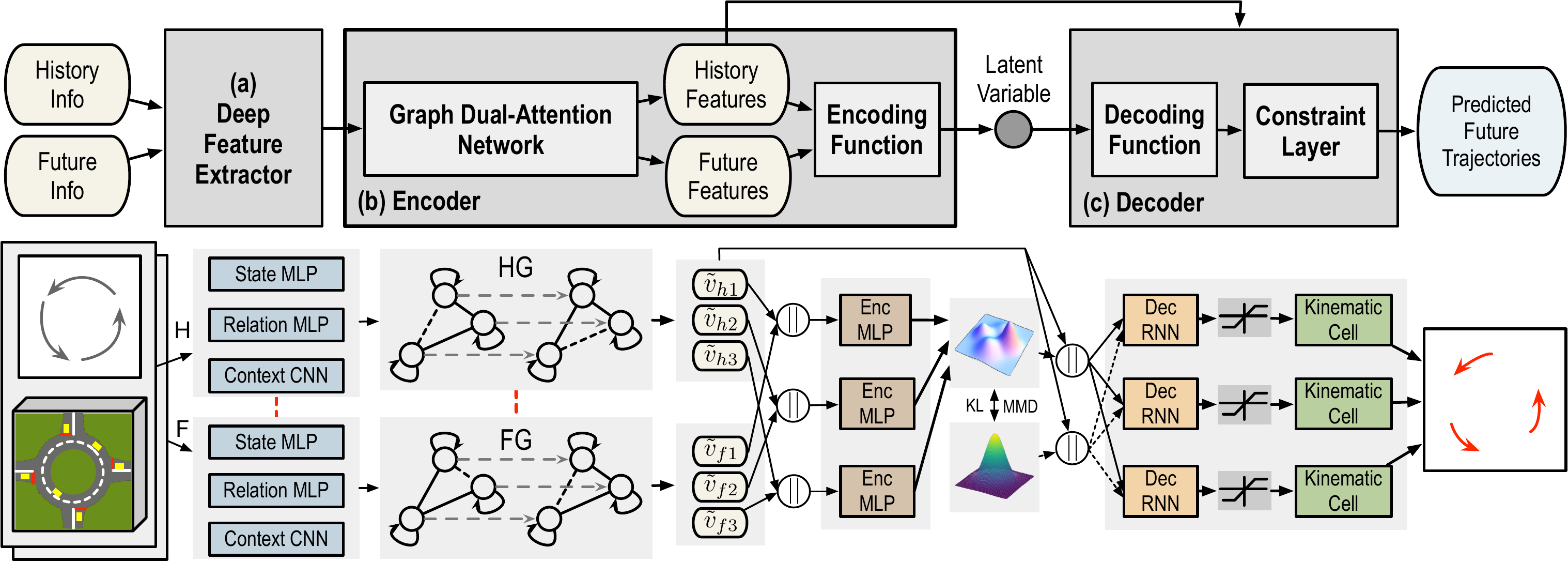}
    \caption{ The detailed architecture of STG-DAT, which consists of three key components: (a) A deep feature extractor which extracts state, relation and context features from the trajectories of agents, the sequences of occupancy density maps and velocity fields. The red dashed lines indicate sharing parameters.
    (b) An encoder which includes a graph dual-attention network that processes spatio-temporal graphs and generates abstract node attributes containing interaction information, and an encoding function which maps the node attributes to a latent space. During the testing phase, the encoding function is not used.
    (c) A decoder which samples future trajectory hypotheses satisfying physical constraints for each agent. 
    The bottom portion of the figure presents some details of (a)-(c).
    $||$ denotes the concatenation operation. MLP refers to multi-layer perceptron. CNN refers to convolutional neural network.}
    %\vspace{-0.3cm}
    \label{fig:model}
\end{figure*}

The detailed architecture of STG-DAT is shown in Fig. \ref{fig:model}, where a standard encoder-decoder architecture is employed.
There are three key components: a deep feature extractor, an encoder with spatio-temporal graph generation and dual-attention network, and a decoder with a kinematic constraint layer.
First, the feature extractor takes in both history and future information and outputs state, relation, and context feature embeddings. 
The information contains the trajectories of the involved interactive agents, and a sequence of context density maps and mean velocity fields.
The scene images or semantic maps can also be included, if they are available in the dataset. Since the neural networks have a the capability of extracting highly flexible features, we choose multi-layer perceptron (MLP) to generate state and relation embeddings and convolutional neural network (CNN) to generate context embedding.
The extracted features are utilized to generate a spatio-temporal graph for both the history and the future, respectively. The node attributes are updated by a spatio-temporal graph attention mechanism.
Then, the updated node attributes are transformed from the feature space into a latent space by an encoding function according to the derivation of the conditional generative modeling. 
Finally, the decoder based on the recurrent neural network generates feasible and human-like future trajectories for all the involved agents.
The number of agents can be flexible in different cases due to the weight sharing and permutation invariance of the graph representation.
All the components are implemented with deep neural networks, thus they can be trained end-to-end efficiently and consistently.

%The detailed architecture of STG-DAT is shown in Fig. \ref{fig:model}, where a standard encoder-decoder architecture is employed.
%There are three key components: a deep feature extractor, an encoder with spatio-temporal graph generation and dual-attention network, and a decoder with a kinematic constraint layer.
%First, the feature extractor takes in both history and future information and outputs state, relation, and context feature embeddings. 
%The information contains the trajectories of the involved interactive agents, and a sequence of context density maps and mean velocity fields.
%The scene images or semantic maps can also be included, if they are available in the dataset.
%The extracted features are utilized to generate a spatio-temporal graph for both the history and the future, respectively. The node attributes are updated by a spatio-temporal graph attention mechanism.
%Then, the updated node attributes are transformed from the feature space into a latent space by an encoding function. Finally, the decoder generates feasible and human-like future trajectories for all the involved agents.
%The number of agents can be flexible in different cases due to the weight sharing and permutation invariance of the graph representation.
%All the components are implemented with deep neural networks, thus they can be trained end-to-end efficiently and consistently.

\subsection{Feature Extraction}
The feature extractor consists of three parts: \textit{State MLP}, \textit{Relation MLP}, and \textit{Context CNN}. The operations below are applied at each time step, and a sequence of state, relation, and context feature embeddings can be obtained.
\begin{itemize}
    \item \textit{State MLP}: It embeds the position, velocity, and heading information into a state feature vector for each agent. Different types of agents use distinct state embedding functions and the same type of agents share the same one. In this paper, we consider three types: vehicles, cyclists and pedestrians. The state embedding (SE) of agent $i$ at time $k$ is obtained by
    \begin{equation}
       SE_i^k = \text{MLP}_s (\bm{\tau}_i^k).
    \end{equation}
    \item \textit{Relation MLP}: It embeds the relative information between each pair of agents into a relation feature vector. We differentiate the edges with opposite directions between the same pair of nodes. The relative information can be either the distance and relative angle (in a 2D polar coordinate), or the differences between the positions of the two agents along two perpendicular axes (in a 2D Cartesian coordinate). We use the latter in this work, since it is simpler to compute and the performance is comparable to the former one.
    More specifically, consider a pair of agents $i$ and $j$. When calculating the relation embedding associated with edge $e_{ij}$ in a Cartesian coordinate, we set agent $i$ as the origin and its heading as the positive direction. The relative position, velocity and heading angle of agent $j$ with respect to agent $i$ can be calculated and denoted as $\bm{\phi}_{ij}^k$.
    The relation embedding (RE) is obtained by 
    \begin{equation}
        RE_{ij}^k = \text{MLP}_r (\bm{\phi}_{ij}^k).
    \end{equation}
    \item \textit{Context CNN}: It extracts spatial features for each agent from a local occupancy density map ($H\times W \times 1$) as well as heuristic features from a local velocity field ($H\times W \times 2$) centered on the corresponding agent. The reason of using occupancy density maps instead of real scene images is to remove redundant information and efficiently represent data-driven drivable regions. 
    This information provides a prior knowledge of common driving behaviors at specific areas of the scene.
    The context embedding (CE) of agent $i$ at time $k$ is obtained by
    \begin{equation}
        CE_i^k = \text{CNN} (\bm{c}_i^k).
    \end{equation}
\end{itemize}

\subsection{Encoder with Graph Dual-Attention Network}
After obtaining the extracted features, a history spatio-temporal graph (HG) and a future spatio-temporal graph (FG) are generated to represent the information related to the involved agents. Here, the state features and context features are concatenated to serve as the (agent) node attributes, whereas the relation features serve as edge attributes. 
The HG and FG contain different time steps, and they are processed in a similar fashion with the graph dual-attention network.
In a specific case, the number of nodes (agents) in both HG and FG is assumed to be fixed, which implies the same agents appear in the whole horizon.
The edges are eliminated at a certain time step if the Euclidean distance between two agents is larger than a threshold $d$. Therefore, the graph connectivity and topology at different time steps may vary.

The proposed graph dual-attention network consists of two consecutive layers: a \textit{topological attention layer} which updates node attributes from the spatial or topological perspective, and a \textit{temporal attention layer} which outputs a high-level feature embedding for each node. The temporal attention layer summaries both the topological and temporal information and figure out relative significance of the information at each time step.
Following the notation in Section III, assume there are totally $n$ nodes (agents) in a graph, we denote a graph as $\mathcal{G}=\{\mathcal{V},\mathcal{E}\}$, where $\mathcal{V}=\{v_i \in \mathbb{R}^{D_n}, i\in \{1,...,n\}\}$ and $\mathcal{E}=\{e_{ij} \in \mathbb{R}^{D_e}, i,j\in \{1,...,n\}\}$. $D_n$ and $D_e$ are the dimensions of node attributes and edge attributes. 

\subsubsection{Topological Attention Layer}

The inputs of this layer are the original spatio-temporal graphs. The output is a new set of node attributes $\Bar{\mathcal{V}}=\{\bar{v}^k_i \in \mathbb{R}^{\bar{D}_n}, i\in \{1,...,n\}, k\in \{1,...,T\}, T=T_h+T_f\}$, which can capture local structural properties. 
The topological attention coefficients $\alpha^k_{ij}$ (showing the significance of node $j$ w.r.t. node $i$) are calculated by
\begin{equation}
%\small
    \alpha^k_{ij} = \frac{\exp{(-A_{ij}(\lambda\norm{v^k_i-v^k_j}^2+\mu\norm{e^k_{ij}}^2))}}{\sum_{p\in N(i)}\exp{(-A_{ip}(\lambda\norm{v^k_i-v^k_p}^2+\mu\norm{e^k_{ip}}^2))}},
\end{equation}
where $N(i)$ is the first-order neighbor nodes (including $i$).
$A_{ij}$ is a prior attention coefficient which provides inductive bias from prior knowledge, $\lambda$ and $\mu$ are weight parameters to adjust the relative importance of node attributes and edge attributes for computing attention coefficients.
The underlying intuition is that the agents, with similar node attributes to the objective agent or with small spatial distance, tend to have more correlation thus should be paid more attention to.
In this work, we set $A_{ij}=1$ implying no prior attention bias, while more exploration on incorporating prior knowledge is left for future work.
Then the node attributes are updated by
\begin{equation}
    \bar{v}^k_i = \sum_{j\in N(i)} f_{\text{act}}(\alpha^k_{ij}W_n v^k_j),
\end{equation}
where $f_\text{act}(\cdot)$ is an activation function, and $W_n$ is learnable parameters. 
The above procedures are applied to each time step, and the weight matrices are shared across different time steps.
We also employ the multi-head attention mechanism \cite{velivckovic2017graph} to boost model performance by adjusting $\lambda$ and $\mu$, where the node attributes obtained by using different attention coefficients are concatenated into a whole vector.
The above message passing procedures can be applied multiple times to capture higher-order interactions with an additional edge update procedure following the form of Eq. (1).
More specifically, the updated edge attributes can be computed by
\begin{equation}
	\bar{e}^k_{ij} = \text{MLP}\left(\left[\bar{v}^k_i, \bar{v}^k_j, e^k_{ij}\right]\right).
\end{equation}

\subsubsection{Temporal Attention Layer}

The input of this layer is the output of the topological attention layer, which is a set of node attributes $\Bar{\mathcal{V}}=\{\bar{v}^k_i \in \mathbb{R}^{\bar{D}_n}, i\in \{1,...,n\}, k\in \{1,...,T\}\}$.
The output is a set of highly abstract node attributes $\widetilde{\mathcal{V}}=\{\widetilde{v}_i \in \mathbb{R}^{\widetilde{D}_n}, i\in \{1,...,n\}\}$. These attributes will be further processed by the downstream modules.
The temporal attention coefficients $\beta^k_i$ is
\begin{equation}
\begin{aligned}
    \beta^{hk}_i =& \frac{\exp{(f_{\text{act}}(\bar{v}^{k \top}_i w))}}{\sum^{T_h}_{k'=1} \exp{(f_{\text{act}}(\bar{v}^{k' \top}_i w))}}, \ (1 \leq k \leq T_h)\\
    \beta^{fk}_i =& \frac{\exp{(f_{\text{act}}(\bar{v}^{k \top}_i w))}}{\sum^{T_h+T_f}_{k'=T_h+1} \exp{(f_{\text{act}}(\bar{v}^{k' \top}_i w))}}, \ (T_h+1 \leq k \leq T_h+T_f)
\end{aligned}
\end{equation}
where $w \in \mathbb{R}^{\bar{D}_n}$ is a weight vector parameterizing the attention function.
Then the node attributes are updated by
\begin{equation}
\begin{aligned}
\widetilde{v}^h_i = \sum^{T_h}_{k=1} f_{\text{act}}(\beta^{hk}_i w^\top \bar{v}^{k}_i), \quad
\widetilde{v}^f_i = \sum^{T_h+T_f}_{k=T_h+1} f_{\text{act}}(\beta^{fk}_i w^\top \bar{v}^{k}_i).
\end{aligned}
\end{equation}
The multi-head attention mechanism can also be employed by learning different $w$ and fusing the information by averaging or concatenation operations.

\subsubsection{Feature Encoding}

For each agent, the history and future node attributes are concatenated and mapped by an encoding function $f_\text{enc}$ to obtain a latent variable $z_i$, which is given by
\begin{equation}
    z_i = f_\text{enc}([\widetilde{v}^h_i||\widetilde{v}^f_i]).
\end{equation}
The underlying intuition is that, during the training phase, the latent variable is able to encode the future information conditioned on the given history information, and it is trained to be consistent with the prior distribution by a regularization term in the loss function.
During the testing phase, although the future information is not available, it can be implicitly obtained by sampling the latent variable from the prior distribution.

\subsection{Decoder with Kinematic Constraint}
\begin{figure}[!tbp]
    \centering
    \includegraphics[width=0.75\columnwidth]{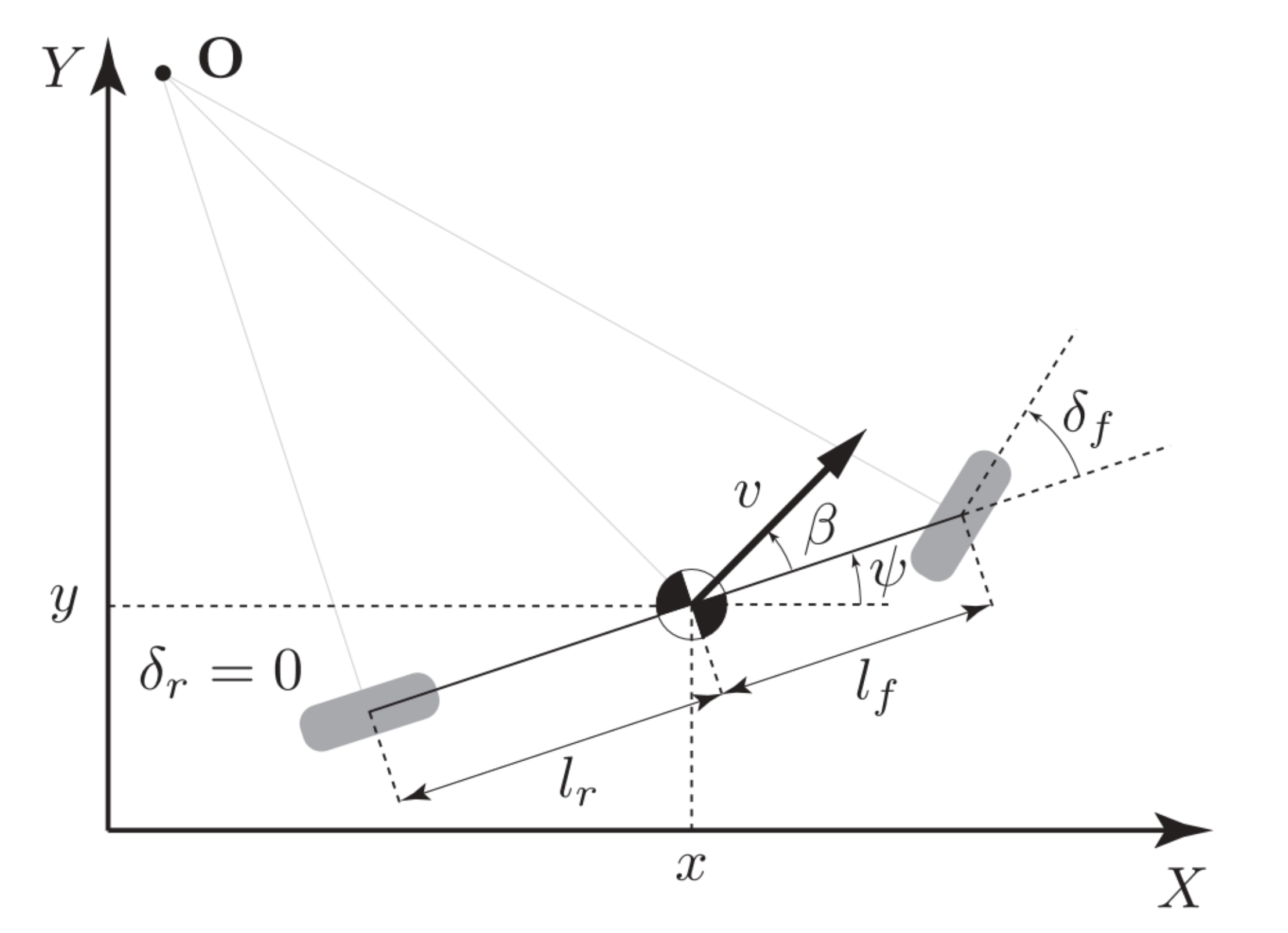}
    \caption{The diagram of the kinematic bicycle model adopted from \cite{kong2015kinematic}. The model equations are provided in Eq. (\ref{eq:discretesystem}).}
    \label{fig:bicycle_model}
    %\vspace{-0.2cm}
\end{figure}

%(b) The detailed unrolled recurrent structure of decoder for a single agent, which consists of a GRU, a saturation function and a kinematics update cell.

We impose a kinematic constraint layer after the decoder gated recurrent unit (GRU) to enforce feasible trajectory prediction.
The same types of agents share the same GRU unit, while different types use distinct ones. This is reasonable since the behavior patterns, speed ranges and traversable areas may be diverse among different types.

The bicycle model is a widely used nonlinear model to approximate the kinematics of vehicles, which is shown in Fig. \ref{fig:bicycle_model}. Here we adopt the discretized form, which is given by
\begin{equation}
\label{eq:discretesystem}
\left\{
	\begin{aligned}
		&x(k+1) = x(k)+v(k) \cos(\psi(k)+\beta(k)) \Delta T,\\
		&y(k+1) = y(k)+v(k) \sin(\psi(k)+\beta(k)) \Delta T,\\
		&\psi(k+1) = \psi(k)+ \dfrac{v(k)}{l_r} \sin \beta(k) \Delta T,\\
		&v(k+1) = v(k)+a(k) \Delta T,\\
		&\beta(k+1) = \beta(k) + \dot{\beta}(k) \Delta T, \\
	\end{aligned}
\right.
\end{equation}
where $(x,y)$ are the coordinates of the center of mass, $\psi$ is the inertial heading and $v$ is the speed of the vehicle. 
$\beta$ is the angle of the current velocity of the center of mass with respect to the longitudinal axis of the car. $l_f$ and $l_r$ denote the distance from the center of mass of the vehicle to the front and rear axles, respectively.
%We denote $\tau_t=[x_t \ y_t \ \psi_t]^\top$, $a_t=[\dot{x}_t \ \dot{y}_t \ \dot{\psi}_t]^\top$ and $u_t=[v_t \ \beta_t]^\top$.
The state of vehicles at time step $k$ is denoted as $\mathbf{s}(k)  = [x(k), y(k), \phi(k), v(k), \beta (k)]^\top$, and the control input at time step $k$ as $ \mathbf{u}(k) = [a(k), \dot{\beta}(k)]^\top $. 
Eq. (\ref{eq:discretesystem}) can be expressed as $ \mathbf{s}(k+1) = \mathbf{f}(\mathbf{s}(k), \mathbf{u}(k))$.

%as shown in Eq. (\ref{KC_equ1}), 
% \begin{equation}
% \label{KC_equ1}
% \begin{aligned}
% \left \{
%     \begin{aligned}
%     &\dot{x} = v\cos{(\psi+\beta)}\\
%     &\dot{y} = v\sin(\psi+\beta)\\
%     &\dot{\psi} = \frac{v}{l_r}\sin(\beta)\\
%     \end{aligned}
% \right.
% \end{aligned}
% \end{equation}

% Then the discretized version of the equations (\ref{KC_equ1}) can be written as 
% \begin{equation}
% \begin{aligned}
% \left \{
% \begin{aligned}
% \tau_{t+1} &= \tau_{t} + a_t \Delta t \\
% u_{t+1} &= u_t + \dot{u}_t \Delta t \\
% a_t &= f(u_t, \dot{u}_t, \tau_t)\\
% \end{aligned}
% \right.
% \end{aligned}
% \end{equation}

The prediction system is expected to provide the position distribution of each agent at each time step. The distribution of the control input $\mathbf{u}(k)$ is assumed to be a multi-variate Gaussian distribution at each time step, which is parameterized by the output of the GRU cell. 
We provide an illustrative example for the agent $i$. The inputs of the gated recurrent unit (GRU) are the node attribute $\widetilde{v}_i$ at the first step and zero paddings for the following steps. The outputs are the raw $\mathbf{u}(k)$ at each step which are truncated by a saturation function in order to restrict the control actions in the feasible range. Then the kinematic cell takes in the current state and action, and outputs the state at the next step.
This procedure is iterated until the prediction horizon is reached.
If $l_r$ is not a prior knowledge or cannot be observed during testing, then we can approximate it with a constant based on the statistics of training data.

In order to propagate the uncertainty to future time steps, two options are available with tradeoffs. The details are introduced in the following. 

\vspace{0.2cm}
\noindent
$\bullet$ \textbf{Nonlinear system with Monte Carlo sampling}:

We can approximate the distribution of position by a set of Monte Carlo samples, which is highly flexible without any restrictions on the nonlinear dynamic system. The particle samples are propagated by the nonlinear bicycle model. The prediction accuracy tends to be improved as the number of particles increases. However, the number of samples may need to be adjusted according to real-time requirements.
If the explicit probability density function of the state variable is required, the kernel density estimation technique can be employed in a non-parametric way.

\vspace{0.2cm}
\noindent
$\bullet$ \textbf{Linearized system with Gaussian assumption}:

Since $\mathbf{u}(k)$ follows Gaussian distribution, we can obtain an analytic distribution of position by linearizing the bicycle model at the current state. 
It is easy to show that the position distribution is also a Gaussian distribution. 
This is simple to implement and computationally efficient, while the flexibility is very limited. This restriction leads to lower prediction accuracy in general, especially when the real distribution is multi-modal.

For a nonlinear system with Gaussian assumption, the expectation and covariance of the state variable can be propagated in a fashion similar to the prior (prediction) update of extended Kalman filter \cite{simon2010kalman}. 
We can linearize the system around the current state $\mathbf{s}(k)$,
\begin{equation}
\label{eq:linearization}
	\mathbf{s}(k+1) = \mathbf{Df}_{s}(k) \mathbf{s}(k) +\mathbf{Df}_{u}(k) \mathbf{u}(k)
\end{equation}
\begin{equation}
\mathbf{u}(k) \sim \mathcal{N}(\bm{\mu}_u(k), \bm{\Sigma}_{uu}(k))
\end{equation}
where $\mathbf{Df}_s(k)$ and $\mathbf{Df}_u(k)$ are the Jacobian matrices defined as below, 
\begin{equation}
\small
\label{eq:Df}
	\begin{aligned}
	&\mathbf{Df}_s(k) = 
	\begin{bmatrix}
	1 & 0 & -\Delta Tv_y(k)  & \Delta Tv_x(k) & -\Delta Tv_y(k)\\
	0 & 1& \Delta Tv_x(k) & \Delta Tv_y(k) & \Delta Tv_x(k)\\
	0 & 0 & 1 & \dfrac{\Delta T}{l_r}\sin\beta(k) & \Delta T \dfrac{v(k)}{l_r} \cos\beta(k)\\
	0 & 0 & 0 & 1 & 0\\
	0 & 0 & 0 & 0 & 1
	\end{bmatrix}, \\
	&\mathbf{Df}_u(k) = 
	\begin{bmatrix}
	0 & 0\\
	0 & 0\\
	0 & 0\\
	\Delta T & 0\\
	0 & \Delta T
	\end{bmatrix},
	\end{aligned}
\end{equation}
where $\gamma(k) = \phi(k) + \beta(k)$, $v_x(k)=v(k)\cos \gamma(k)$, and $v_y(k)=v(k)\sin \gamma(k)$.
Then we have the distribution of $\mathbf{s}(k+1)$, which is expressed as
\begin{equation}
	\mathbf{s}(k+1) \sim \mathcal{N}(\bm{\mu}_s(k+1), \bm{\Sigma}_{ss}(k+1)),
\end{equation}
where 
\begin{equation}
\small
	\label{eq:distribution}
	\begin{aligned}
		&\bm{\mu}_s(k+1) = \mathbf{Df}_s(k) \bm{\mu}_s(k) + \mathbf{Df}_u(k) \bm{\mu}_u(k), \\
		&\bm{\Sigma}_{ss}(k+1) = \mathbf{Df}_s(k)\bm{\Sigma}_{ss}(k)\mathbf{Df}_s(k)^\top+\mathbf{Df}_u(k)\bm{\Sigma}_{uu}(k)\mathbf{Df}_u(k)^\top,
	\end{aligned}
\end{equation}
with the initial condition $ \mathbf{s}(0) \sim \mathcal{N}(\bm{\mu}_s(0), \bm{\Sigma}_{ss}(0))$.

\begin{figure}[!tbp]
    \centering
    \includegraphics[width=\columnwidth]{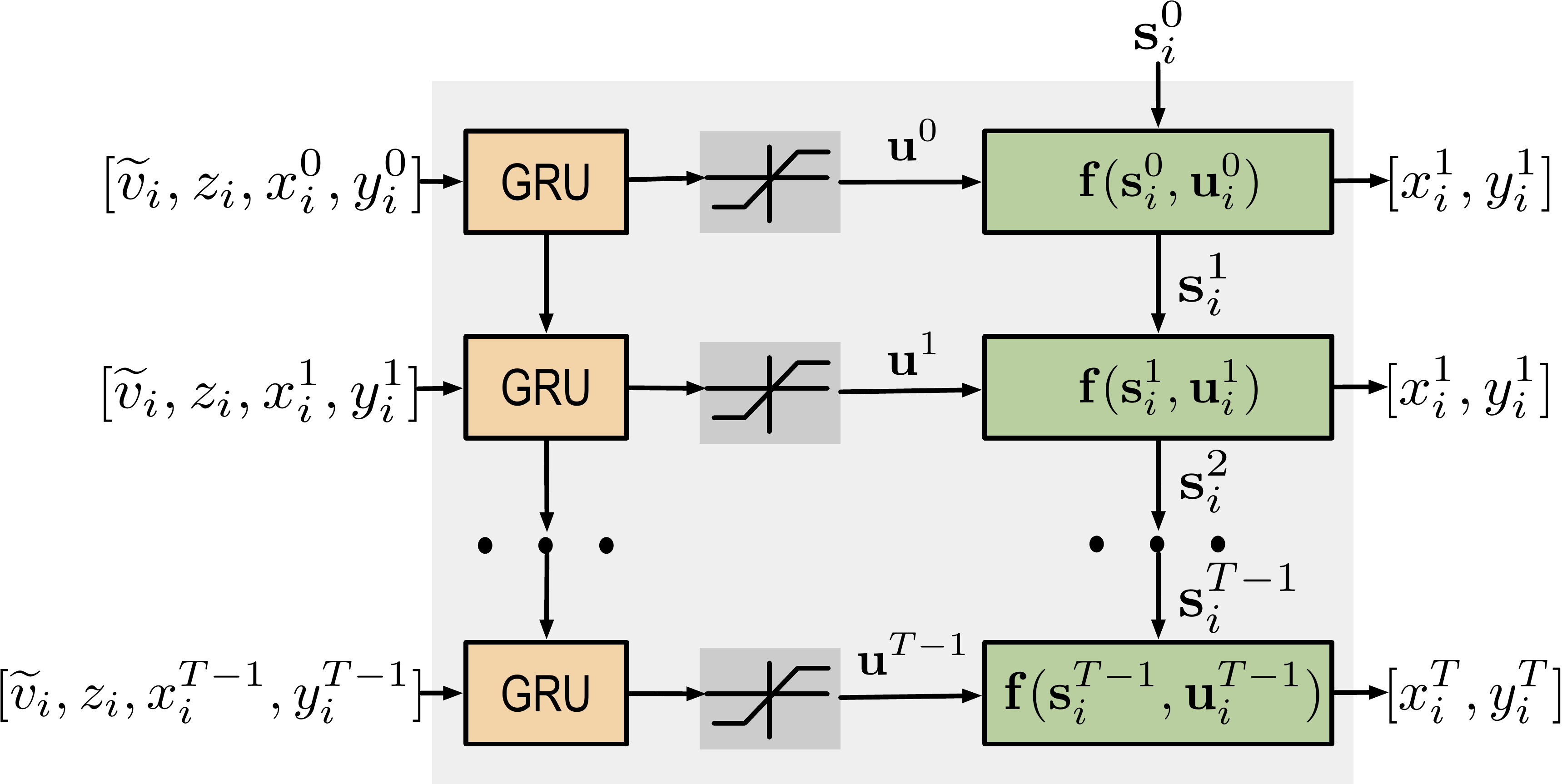}
    \caption{The diagram of the recurrent decoder with kinematic constraint layer. The recurrent process consists of two phases: burn-in phase and prediction phase. In the burn-in phase, the history groundtruth is used as the input of GRU at each step for initialization purpose. In the prediction phase, the output position at the last step will serve as the input of the next step. The iteration continues until the prediction horizon is reached.}
    \label{fig:decoder}
\end{figure}

\subsection{Loss Function and Training}
In this part, we demonstrate the loss function of our model, which is based on the optimization formulation of Wasserstein generative modeling aforementioned in Section III.
In order to keep consistent with Section III, we use the same notations: $x$ denotes the predicted trajectories, $z$ denotes the latent variable, and $y$ denotes the condition variable. 
%These notations were used for other purposes in previous sections.

The optimization problem can be formulated in the same way as in \cite{ma2019}, which is written as
\begin{equation}
\begin{aligned}
\min_{\theta,\phi, s.t. 0<1-\alpha<\beta} &-\mathbf{E}_{p_{\phi}(z|x, y)}[\log p_{\theta}(x|z, y)]]\\
&+\alpha\mathbf{E}_{p(x|y)}[D_{\text{KL}}[p_{\phi}(z|x, y)||p(z|y)]]\\
&+\beta D(p_{\phi}(z|y), p(z|y)).\\
\end{aligned}
\end{equation}
The detailed derivation can be found in \cite{ma2019}.
Since the whole system is fully differentiable, we can train the network in an end-to-end fashion by the Adam optimizer \cite{kingma2014adam}. The loss function is given by
\begin{equation}
    \begin{aligned}
    \mathcal{L} =\ & \gamma  \mathbf{E}_{j \in \{1,...,N_b\}} \norm{\mathbf{\tau}_j^{T_h+1:T_h+T_f} - \hat{\mathbf{\tau}}_{j}^{T_h+1:T_h+T_f}}^2 \\ 
    + \ &\alpha  \mathbf{E}_{p(x|y)}[D_{\text{KL}}[p_{\phi}(z|x,y)||p(z|y)] \\
    + \ &\beta  \text{MMD}(p_{\phi}(z|y), p(z|y)),
    \end{aligned}
\end{equation}
where 
\begin{equation}
    \centering
    \begin{aligned}
    p_{\phi}&(z|x, y) = \mathcal{N}(\text{MLP}(\tilde{v}_h, \tilde{v}_f), \mathbf{I}), \\
    \tilde{v}_h &= \text{GDAT}(\text{FE}(y)), \ \tilde{v}_f = \text{GDAT}(\text{FE}(x)),\\
    y &= \{\mathbf{T}_{1:T_h}, \mathbf{C}_{1:T_h}\}, \\
    x &= \{\mathbf{T}_{T_h+1:T_h+T_f}, \mathbf{C}_{T_h+1:T_h+T_f}\}.
    \end{aligned}
\end{equation}
$\text{FE}$ is the deep feature extractor and $\text{GDAT}$ is the proposed graph dual-attention network.
$\gamma$ is a weight parameter to adjust the relative importance of the reconstruction loss, $N_b$ is the total number of training agents, $D_\text{KL}$ is Kullback-Leibler divergence, and MMD is maximum mean discrepancy. If $\gamma \gg \alpha,\beta$, then the loss function degenerates to the mean squared error loss.
The whole model is trained in an end-to-end fashion.

\begin{table*}[htbp]
	%\vspace{-0.5cm}
	\caption{ADE / FDE (meters) Comparisons (ETH \& UCY datasets). }
	\fontsize{8}{8}\selectfont
	\resizebox{\textwidth}{!}{
		%\begin{center}
		\begin{tabular}{m{1.8cm}<{\centering}| m{1.8cm}<{\centering}| m{1.8cm}<{\centering}| m{1.8cm}<{\centering}| m{1.8cm}<{\centering}| m{1.8cm}<{\centering}|| m{1.8cm}<{\centering} }
			\toprule
			\midrule
			Scenes & S-LSTM & S-GAN & CGNS & Social-BiGAT & Trajectron & \textbf{STG-DAT}\\ % \hhline{=|=|=}
			\midrule 
			ETH    &  1.09 / 2.35  &  0.81 / 1.52  &  0.62 / 1.40  &  0.69 / 1.29  &  0.48 / 0.93  &  \textbf{0.38} / \textbf{0.77} \\ 
			HOTEL  &  0.79 / 1.76  &  0.72 / 0.61  &  0.70 / 0.93  &  0.49 / 1.01  &  0.29 / 0.54  &  \textbf{0.25} / \textbf{0.39} \\ 
			UNIV   &  0.67 / 1.40  &  0.60 / 1.26  &  0.48 / 1.22  &  0.55 / 1.32  &  0.44 / 0.93  &  \textbf{0.41} / \textbf{0.82} \\ 
			ZARA1  &  0.47 / 1.00  &  0.34 / 0.69  &  0.32 / 0.59  &  0.30 / 0.63  &  0.35 / 0.68  &  \textbf{0.23} / \textbf{0.50} \\ 
			ZARA2  &  0.56 / 1.17  &  0.42 / 0.84  &  0.35 / 0.71  &  0.36 / 0.75  &  0.36 / 0.70  &  \textbf{0.21} / \textbf{0.46} \\ 
			\midrule
			AVG    &  0.72 / 1.54  &  0.58 / 1.18  &  0.49 / 0.97  &  0.48 / 1.00  &  0.46 / 0.94   & \textbf{0.30} / \textbf{0.59} \\ 
			\bottomrule
		\end{tabular}
		%\end{center}
	}
	\label{tab:ETH_UCY}
	%\vspace{-0.3cm}
\end{table*}

\begin{table*}[htbp]
	%\vspace{-0.5cm}
	\caption{ADE / FDE (pixels) Comparisons (SDD dataset).}
	\fontsize{8}{8}\selectfont
	\resizebox{\textwidth}{!}{
		%\begin{center}
		\begin{tabular}{m{1.8cm}<{\centering}| m{1.8cm}<{\centering}| m{1.8cm}<{\centering}| m{1.8cm}<{\centering}| m{1.8cm}<{\centering}|| m{1.8cm}<{\centering}| m{1.8cm}<{\centering} }
			\toprule
			\midrule
			S-LSTM & S-GAN  & CGNS & DESIRE & Trajectron & \textbf{STG-DAT} (same node) &\textbf{STG-DAT}\\ % \hhline{=|=|=}
			\midrule 
			33.19 / 56.38  & 24.81 / 38.62 & 15.60 / 28.20 &  19.30 / 34.12 & 17.38 / 31.46 & 14.55 / 23.54 & \textbf{13.25} / \textbf{21.94} \\ 
			\bottomrule
		\end{tabular}
		%\end{center}
	}
	\label{tab:SDD}
	%\vspace{-0.1cm}
\end{table*}

\section{Experiments}
In this section, we validate the proposed method on three publicly available benchmark datasets for trajectory prediction of traffic participants. The results are analyzed and compared with state-of-the-art baselines.

\subsection{Datasets}
Here we briefly introduce the datasets below. Please refer to the appendix for data processing details.

$\bullet$ \textbf{ETH} \cite{ETH} \textbf{and UCY} \cite{UCY}: 
These two datasets are usually used together in literature, which include bird-eye-view videos and annotations of pedestrians in both indoor and outdoor scenarios. The trajectories were extracted in the world space (unit: meters).

$\bullet$ \textbf{Stanford Drone Dataset (SDD)}\cite{SDD}:
The dataset also contains a set of bird-eye-view images and the corresponding trajectories of involved entities. It was collected in multiple scenarios in a university campus full of interactive pedestrians, cyclists and vehicles. The trajectories were extracted in the image pixel space.

$\bullet$ \textbf{INTERACTION Dataset (ID)}\cite{interactiondataset}:
The dataset contains naturalistic motions of various traffic participants in a variety of highly interactive driving scenarios. Trajectory data was collected using drones and traffic cameras. 
The high-definition maps of scenarios and agents' trajectories are provided. We consider three types of scenarios: roundabout (RA), unsignalized intersection (UI) and highway ramp (HR). The trajectories were extracted in the world space (unit: meters).

\subsection{Evaluation Metrics}
We evaluate the model performance in terms of average displacement error (ADE) and final displacement error (FDE), which are exactly the same as \cite{alahi2016social,gupta2018social,kosaraju2019social}.
ADE is defined as the average distance between the predicted trajectories and the groundtruth over all the involved entities within the prediction horizon.
FDE is defined as the deviated distance at the last predicted time step. 
For the ETH, UCY, and SDD dataset, 
we predicted the future 12 time steps (4.8s) based on the historical 8 time steps (3.2s).
For the ID dataset, we predicted the future 10 time steps (5.0s) based on the historical 4 time steps (2.0s).

% We compared the performance of our proposed method with the following baseline approaches: Constant Velocity Model (CVM), Probabilistic LSTM (P-LSTM) \cite{Jiachen_ICRA19}, Social Forces (S-Forces) \cite{luber2010people}, Social LSTM (S-LSTM) \cite{alahi2016social}, Social GAN \cite{gupta2018social}, Social Attention (S-ATT) \cite{vemula2018social}, DESIRE \cite{lee2017desire}, Social-BiGAT \cite{kosaraju2019social} and Trajectron \cite{ivanovic2019trajectron}. Please refer to the reference papers for more details.

\subsection{Baseline Methods}

\begin{itemize}
    %\item \textbf{Constant Velocity Model (CVM)}: This is a widely used linear vehicle kinematics model with a constant velocity assumption. This model is also able to forecast trajectories of pedestrians. A Gaussian noise term is injected at each time step to model uncertainty.
    %\item \textbf{Linear Regression (LR)}: A linear predictor which minimizes the least square error over the historical trajectories.
    \item \textbf{Probabilistic LSTM (P-LSTM)} \cite{li2019interaction}: The backbone of the model is the same as an encoder-decoder architecture with vanilla LSTM. In order to incorporate uncertainty in the model, a noise term sampled from the normal distribution is added in the input, which results in a probabilistic model.
    %\item \textbf{Social-Forces} \cite{luber2010people}: The model is based on the concepts developed in the cognitive and social science communities that describe individual and collective pedestrian dynamics.
    \item \textbf{Social LSTM (S-LSTM)} \cite{alahi2016social}: The trajectories are encoded with an LSTM layer, whose hidden states serve as the input of the proposed social pooling layer, which handles interaction modeling implicitly. 
    \item \textbf{Social GAN (S-GAN)}  \cite{gupta2018social}: The model introduces a generative adversarial learning scheme into S-LSTM to further improve performance.
    \item \textbf{Social Attention (S-ATT)} \cite{vemula2018social}: The model is based on the architecture of Structural-RNN \cite{jain2016structural}, which deals with spatio-temporal graphs with recurrent neural networks.
    %\item \textbf{Clairvoyant attentive recurrent network (CAR-Net)} \cite{sadeghian2018car}: The model employs a physical attention module to capture agent-space interaction but without considering interactions among agents.
    %\item \textbf{Social-Ways} \cite{amirian2019social}: The model uses a generative adversarial network (Info-GAN) to sample plausible predictions for any agent in the scene. 
    %\item \textbf{SoPhie} \cite{sadeghian2019sophie}: The model leverages both context images and trajectory information to generate paths compliant to social and physical constraints.
    \item \textbf{DESIRE} \cite{lee2017desire}: The model is a deep stochastic inverse optimal control framework based on conditional variational auto-encoder with RNN encoders and decoders. A ranking module for sampled trajectories was introduced to indicate their likelihood.
    \item \textbf{Social-BiGAT} \cite{kosaraju2019social}: The model is a graph-based generative adversarial network, which is based on a graph attention network. A recurrent encoder-decoder architecture is trained via an adversarial scheme.
    \item \textbf{Trajectron} \cite{ivanovic2019trajectron}: The model combines tools from recurrent sequence modeling and variational deep generative modeling to produce a distribution of future trajectories.
\end{itemize}

%%%%%%%%%%%%%%%%%%%%%%%%%%%%%%%%%%%%
\begin{table*}[htbp]
	%\vspace{-0.5cm}
	\caption{ADE / FDE (meters) Comparisons (ID dataset).}
	\fontsize{8}{8}\selectfont
	\begin{center}
		\resizebox{0.95\textwidth}{!}{
			\begin{tabular}{m{0.8cm}<{\centering}|m{1cm}<{\centering}|m{1.8cm}<{\centering}| m{1.8cm}<{\centering}| m{1.8cm}<{\centering}| m{1.8cm}<{\centering}| m{1.8cm}<{\centering}| m{1.8cm}<{\centering}}
				\toprule
				\midrule
				\multirow{4}*{\shortstack[lb]{}} 
				&	& \multicolumn{6}{c}{Baseline Methods} \\
				\cline{3-8}
				& & & & & & &\\[-0.1cm]
				Scenes & Time  & P-LSTM & S-LSTM & S-GAN & S-ATT & CGNS & Trajectron \\[-0.1cm]% \hhline{=|=|=}
				& & & & & & &\\
				\midrule 
				\multirow{5}*{RA}  
				&1.0s	 &  0.13 / 0.15  &   0.16 / 0.20  &  0.13 / 0.16  &  0.14 / 0.17  & 0.11 / 0.16 & 0.09 / 0.13\\ 
				&2.0s    &  0.27 / 0.35  &   0.25 / 0.39  &  0.22 / 0.35  &  0.23 / 0.39  & 0.21 / 0.32 & 0.18 / 0.30\\ 
				&3.0s	 &  0.58 / 0.83  &   0.52 / 0.80  &  0.45 / 0.72  &  0.53 / 0.77  & 0.44 / 0.68 & 0.40 / 0.61\\ 
				&4.0s	 &  0.87 / 1.35  &   0.84 / 1.33  &  0.70 / 1.19  &  0.78 / 1.25  & 0.69 / 1.03 & 0.64 / 0.98\\ 
				&5.0s    &  1.36 / 1.88  &   1.29 / 1.80  &  1.13 / 1.51  &  1.22 / 1.69  & 1.01 / 1.45 & 0.95 / 1.32\\ 
				\midrule
				\multirow{5}*{UI}  
				&1.0s	 &  0.11 / 0.20  &   0.12 / 0.17  &  0.11 / 0.16  &  0.12 / 0.18  & 0.11 / 0.14 & 0.10 / 0.14\\ 
				&2.0s	 &  0.28 / 0.49  &   0.25 / 0.48  &  0.24 / 0.40  &  0.26 / 0.43  & 0.24 / 0.38 & 0.23 / 0.35\\ 
				&3.0s	 &  0.45 / 0.88  &   0.41 / 0.83  &  0.39 / 0.77  &  0.41 / 0.80  & 0.38 / 0.76 & 0.36 / 0.72\\ 
				&4.0s	 &  0.83 / 1.64  &   0.77 / 1.47  &  0.74 / 1.31  &  0.75 / 1.39  & 0.70 / 1.18 & 0.69 / 1.10\\ 
				&5.0s    &  1.34 / 2.36  &   1.31 / 2.20  &  1.21 / 1.98  &  1.29 / 2.06  & 1.06 / 1.90 & 1.01 / 1.84\\ 
				\midrule
				\multirow{5}*{HR}  
				&1.0s	 &  0.07 / 0.11  &   0.06 / 0.10  &  0.06 / 0.08  &  0.06 / 0.09  & 0.05 / 0.08 & 0.05 / 0.07 \\ 
				&2.0s	 &  0.21 / 0.39  &   0.19 / 0.35  &  0.17 / 0.31  &  0.17 / 0.32  & 0.16 / 0.29 & 0.15 / 0.27\\ 
				&3.0s	 &  0.55 / 1.02  &   0.49 / 0.92  &  0.43 / 0.81  &  0.45 / 0.84  & 0.40 / 0.75 & 0.38 / 0.70\\ 
				&4.0s	 &  0.90 / 1.58  &   0.81 / 1.43  &  0.71 / 1.26  &  0.74 / 1.31  & 0.66 / 1.17 & 0.62 / 1.09\\ 
				&5.0s    &  1.51 / 2.57  &   1.36 / 2.31  &  1.20 / 2.04  &  1.25 / 2.12  & 1.12 / 1.90 & 1.04 / 1.77\\ 
				\bottomrule
			\end{tabular}
		}
	\end{center}
	\label{tab:ID1}
	%\vspace{-0.cm}
\end{table*}
\begin{table*}[htbp]
	%\vspace{-0.8cm}
	\fontsize{7.5}{8}\selectfont
	\begin{center}
		\resizebox{0.95\textwidth}{!}{
			\begin{tabular}{m{0.7cm}<{\centering}|m{0.9cm}<{\centering}|m{2.7cm}<{\centering}| m{2.7cm}<{\centering}| m{2.7cm}<{\centering}| m{2.7cm}<{\centering}}
				\toprule
				\midrule
				\multirow{4}*{\shortstack[lb]{}} 
				&	& \multicolumn{4}{c}{\textbf{STG-DAT}} \\
				\cline{3-6}
				& & & & & \\%[-0.1cm]
				Scenes & Time  & $\mathbf{T}$  & $\mathbf{T}+\mathbf{C}-\mathbf{ATT}$ & $\mathbf{T}+\mathbf{C}$ &  $\mathbf{T}+\mathbf{C}+\textbf{\text{K}}$ \\[-0.1cm]% \hhline{=|=|=}
				& & & & & \\
				\midrule 
				\multirow{5}*{RA}  
				&1.0s 	& 0.07 / 0.11 & 0.09 / 0.13 & 0.08 / 0.13 & \textbf{0.06} / \textbf{0.10}   \\ 
				&2.0s 	& 0.19 / 0.31 & 0.22 / 0.34 & 0.16 / 0.29 & \textbf{0.14} / \textbf{0.25}  \\ 
				&3.0s   & 0.35 / 0.57 & 0.43 / 0.66 & 0.31 / 0.54 & \textbf{0.26} / \textbf{0.47}   \\ 
				&4.0s   & 0.58 / 0.92 & 0.70 / 1.05 & 0.51 / 0.83 & \textbf{0.42} / \textbf{0.71}  \\ 
				&5.0s   & 0.90 / 1.25 & 1.08 / 1.56 & 0.85 / 1.17 & \textbf{0.68} / \textbf{1.01}   \\ 
				\midrule
				\multirow{5}*{UI}  
				&1.0s 	& \textbf{0.06} / 0.10 & 0.09 / 0.16 & 0.08 / 0.14 & 0.07 / \textbf{0.11} \\ 
				&2.0s 	& 0.20 / 0.32 & 0.23 / 0.36 & 0.19 / 0.30 & \textbf{0.16} / \textbf{0.28} \\ 
				&3.0s 	& 0.37 / 0.65 & 0.41 / 0.71 & 0.34 / 0.59 & \textbf{0.30} / \textbf{0.55} \\ 
				&4.0s   & 0.60 / 0.98 & 0.72 / 1.19 & 0.55 / 0.91 & \textbf{0.49} / \textbf{0.83} \\ 
				&5.0s   & 0.97 / 1.69 & 1.22 / 1.91 & 0.88 / 1.50 & \textbf{0.77} / \textbf{1.26} \\ 
				\midrule
				\multirow{5}*{HR}  
				&1.0s 	& \textbf{0.04} / 0.07 & 0.05 / 0.07 & \textbf{0.04} / 0.07 & \textbf{0.04} / \textbf{0.06}  \\ 
				&2.0s 	& 0.14 / 0.25 & 0.15 / 0.27 & 0.13 / 0.24 & \textbf{0.12} / \textbf{0.22} \\ 
				&3.0s 	& 0.35 / 0.66 & 0.37 / 0.70 & 0.34 / 0.64 & \textbf{0.31} / \textbf{0.58} \\ 
				&4.0s 	& 0.58 / 1.02 & 0.61 / 1.08 & 0.56 / 0.99 & \textbf{0.51} / \textbf{0.90} \\ 
				&5.0s   & 0.97 / 1.65 & 1.03 / 1.75 & 0.95 / 1.61 & \textbf{0.86} / \textbf{1.46} \\ 
				\bottomrule
			\end{tabular}
			
		}
	\end{center}
	\label{tab:ID2}
	%\vspace{-0.cm}
\end{table*}

\subsection{Implementation Details}
A batch size of 64 was used and the models were trained for 100 epochs with early stopping using the Adam optimizer with an initial learning rate of 0.001. The models were trained on a single NVIDIA TITAN X GPU. We used a split of 70\%, 10\%, 20\% as training, validation and testing data, respectively.
During the testing phase, the data processing time for a single time step is around 8ms in average. According to our setting, we predict the future 10 time steps which needs around 80ms (equivalent to 12.5Hz).

The details of our model architecture are introduced in the following.
\begin{itemize}
    \item \textbf{Deep Feature Extractor (FE)}: The \textit{State MLP} and \textit{Relation MLP} both have three hidden layers with 128 hidden units. The \textit{Context CNN} adopts the same backbone structure of ResNet18 \cite{he2016deep}, which is trained from scratch.
    \item \textbf{Graph Dual-Attention Network (GDAT)}: The dimensions of node attributes and edge attributes are 64 and 16, respectively. These dimensions are fixed in different rounds of message passing. The activation functions in the attention mechanism are LeakyReLU.
    \item \textbf{Encoding Function}: The encoding function is a three-layer MLP with 128 hidden units. The dimension of latent variable is 32.
    \item \textbf{Decoding Function}: The decoding function is a GRU recurrent layer with 128 hidden units.
\end{itemize}

\subsection{Quantitative Analysis}
%We analyze the prediction results quantitatively in the following.

$\bullet$ \textbf{ETH and UCY Datasets}:
The comparison of the proposed STG-DAT and baseline methods in terms of ADE and FDE is shown in Table \ref{tab:ETH_UCY}. Some of the reported statistics are adopted from the original papers.
%It is not surprising that the linear model performs the worst in general since it does not consider any social interactions or context information. An exception is the HOTEL scenario since most trajectories are relatively straight and can be well approximated by line segments.
%The P-LSTM is able to achieve smaller prediction error than LR due to the larger model capacity and flexibility of recurrent neural network, although they both predict solely based on the individual's historical trajectories.
All the baseline methods deal with interaction modeling in a specific way.
The S-LSTM employs a social pooling mechanism to model the interactions between entities.
%The S-ATT uses spatial attention mechanism to figure out relative significance.
The S-GAN improve the performance by introducing deep generative modeling.
The CGNS combines conditional latent space learning and variational divergence minimization to further enhance the generation capability. 
The Trajectron adopts a graph-structured model with sequence modeling.
Both Social-BiGAT and our method leverage the trajectory and context information, but in different ways. Our model can achieve better performance owing to the explicit interaction modeling with graph neural networks and more compact distribution learning with conditional Wasserstein generative modeling. 
In general, our approach achieves the smallest ADE and FDE across different scenes. 
The average ADE / FDE are reduced by 34.8\% / 37.2\% compared to the best baseline (Trajectron).

$\bullet$ \textbf{Stanford Drone Dataset}: 
The comparison of results is provided in Table \ref{tab:SDD}, where the ADE and FDE are reported in the pixel distance.
Note that we also included cyclists and vehicles in the dataset besides pedestrians.
%Similarly, the P-LSTM performs the worst due to lack of relational reasoning. 
%The S-Forces incorporates interaction modeling from an energy-based perspective, while the S-ATT utilizes attention mechanisms, which leads to better accuracy.
In general, the relative performance of baseline methods is consistent with the observation on the ETH / UCY datasets.
%The S-GAN and DESIRE both solve the task from a probabilistic perspective by learning implicit data distribution and latent space representations, respectively.
Our approach achieves the best performance in terms of prediction error, which implies the superiority of explicit interaction modeling and necessity of leveraging both trajectory and context information. 
In order to show the effectiveness of distinct node embedding functions for different types, we provide an ablative result by treating all the agents as the same type, which leads to an increase on the prediction error.
For the full model, the ADE / FDE are reduced by 15.1\% / 22.4\% with respect to the best baseline (CGNS).

$\bullet$ \textbf{INTERACTION Dataset}: 
We finally compare the model performance on the real-world driving dataset in Table \ref{tab:ID1}. 
For fair comparison, we only involved the baseline approaches whose codes are publicly available and can be adapted to the same setup as our approach. 
Although we trained a unified prediction model on different scenarios simultaneously, we analyzed the results for each type of scenario separately.
In the HR scenarios, the results of baseline methods are comparable while our model achieves the best performance. The behavior patterns of vehicles in HR scenarios are relatively easy to forecast, since most vehicles are doing car following without highly interactive behaviors. 
In the RA and UI scenarios, however, the superiority of the proposed system is more distinguishable due to frequent interactions.
The P-LSTM performs the worst since it predicts future trajectories for each agent individually without considering their relations.
Although all the other baseline approaches incorporate interaction modeling by different strategies which further reduce the prediction error, our model still performs the best. This implies the advantages of graph dual-attention network for interaction modeling, as well as kinematic layer for feasibility constraints.
By using our full model $\mathbf{T}+\mathbf{C}+\mathbf{K}$, the 5.0s ADE / FDE are reduced by 28.4\% / 23.5\%, 33.7\% / 31.5\% and 17.3\% / 17.5\% in RA, UI and HR with respect to the best baseline (Trajectron), respectively.

We also tested the tracking performance using our approach and baseline methods, which is shown in Table IV.
The constant velocity model (CVM) and constant acceleration model (CAM) are widely used linear vehicle kinematics models with a constant velocity / acceleration assumption. These models are employed frequently in multi-target tracking literature. A Gaussian noise term is injected at each time step to model uncertainty.
It shows that tracking with learning-based models performs consistently better than CVM and CAM due to the ability of interaction-aware prediction.
Our approach STG-DAT achieves a significantly higher accuracy.

\subsection{Qualitative and Ablative Analysis}
\begin{figure*}[!tbp]
    \centering
    \includegraphics[width=\textwidth]{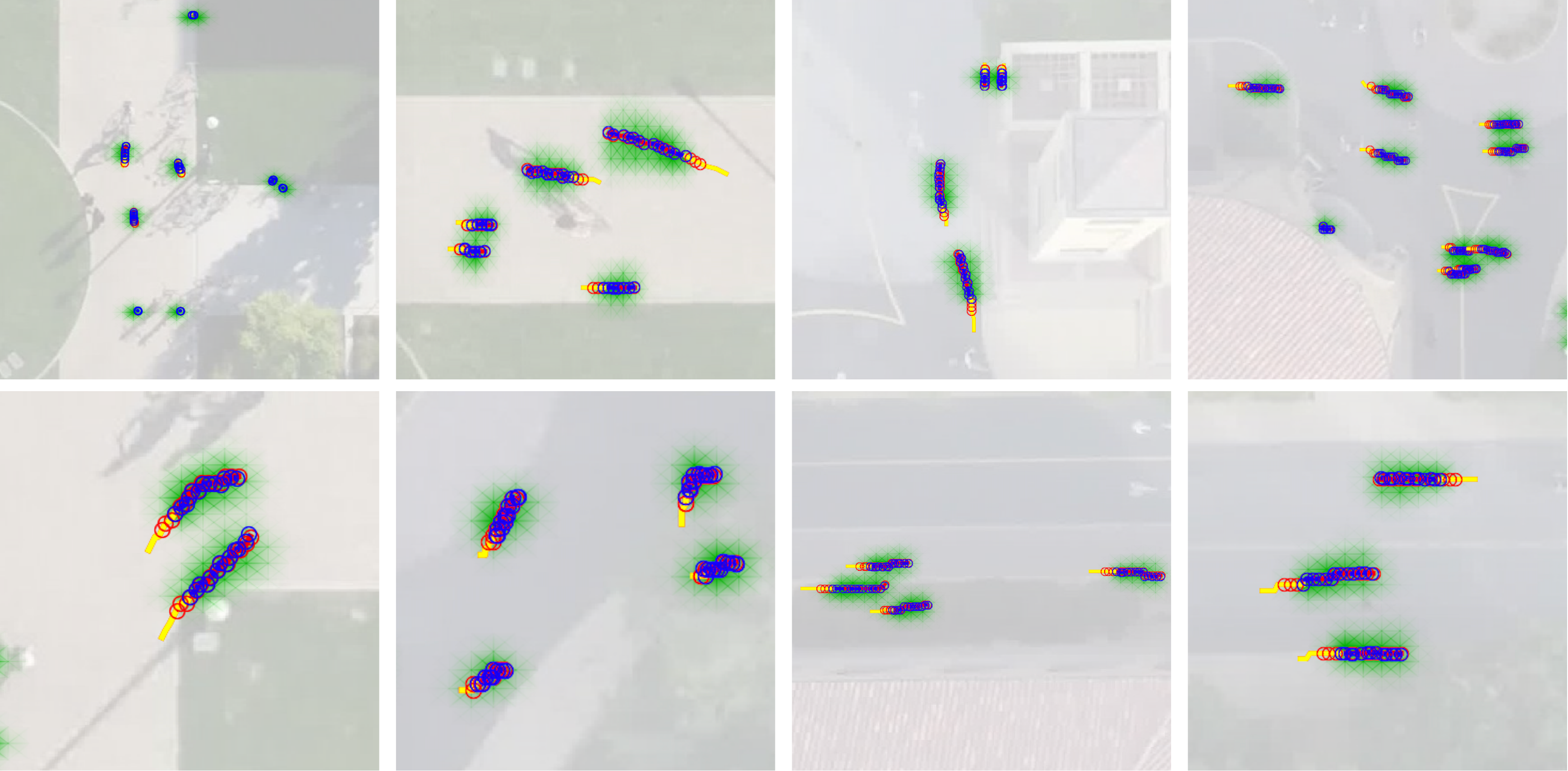}
    \caption{Qualitative results on the SDD dataset. The green mask represents the predicted distribution and the yellow, blue and red lines represent historical observation, groundtruth and a trajectory hypothesis sampled from the distribution with the smallest error, respectively.
    }
    \label{fig:sdd_plot}
    %\vspace{-0.5cm}
\end{figure*} 

\begin{figure*}[!tbp]
    \centering
    \includegraphics[width=\textwidth]{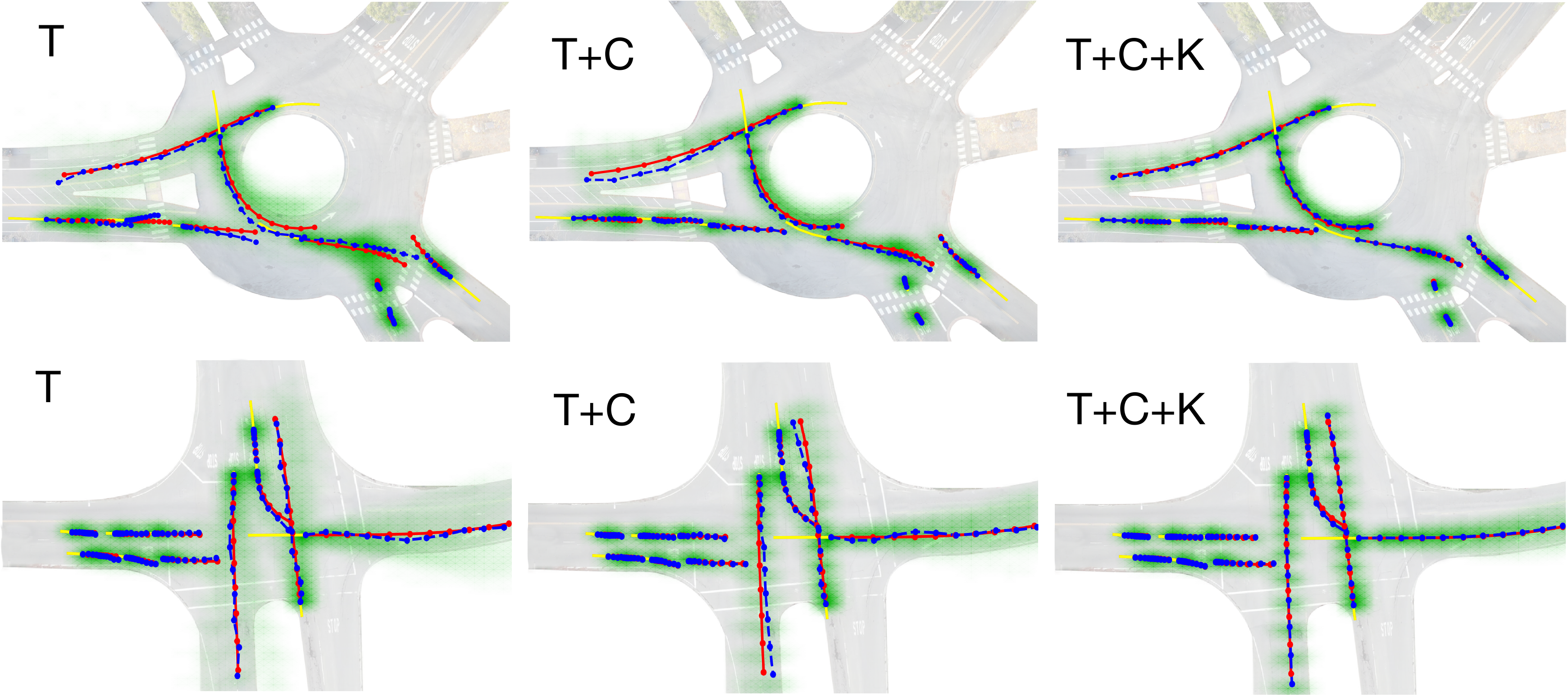}
    \caption{Qualitative and ablative results on the ID dataset. The green mask represents the predicted distribution and the yellow, blue and red lines represent historical observation, groundtruth and a trajectory hypothesis sampled from the distribution with the smallest error, respectively.
    }
    \label{fig:ID_plot}
    %\vspace{-0.5cm}
\end{figure*} 

\begin{table}[!tbp]
	\centering  
	\caption{Tracking Performance of Vehicle Positions and Velocities}  
	\begin{tabular}{m{3.5cm}<{\centering}| m{2.cm}<{\centering} | m{2.cm}<{\centering} }  
		\toprule  
		\midrule
		Method & Position (m) & Velocity (m/s) \\    
		\midrule        
		CVM  			& 0.025  & 0.231  \\  
		CAM  			& 0.021  & 0.186 \\  
		\midrule
		P-LSTM  		& 0.014  & 0.108\\  
		S-LSTM  		& 0.013  & 0.101  \\  
		S-GAN  			& 0.011  & 0.087 \\  
		S-ATT  			& 0.012  & 0.096 \\  
		CGNS  			& 0.010  & 0.077  \\ 
		Trajectron 		& 0.009  & 0.061           \\  
		\midrule
		STG-DAT (Linearization)  &   0.007  &  0.035\\
		\textbf{STG-DAT (Monte Carlo)}    &   \textbf{0.005}  &  \textbf{0.030}\\
		\bottomrule  
	\end{tabular}  
\end{table}

We qualitatively evaluated on prediction hypotheses of typical testing cases on the SDD dataset and ID dataset in Fig. \ref{fig:sdd_plot} and Fig. \ref{fig:ID_plot}, respectively. 
Although we jointly predict all the agents in a scene, we show predictions for a subset for clearness.
It shows that our approach can handle different challenging scenarios (e.g. intersection, roundabout) and diverse behaviors (e.g. going straight, turning, waiting, stopping) of vehicles and pedestrians. 
Generally, the groundtruth trajectories are close to the mean of predicted distribution and the model also allows for uncertainty.

We also conducted comprehensive ablative analysis on the ID dataset to demonstrate relative significance of context information, dual-attention mechanism and the kinematic constraint layer for vehicle trajectory prediction. 
The descriptions of compared model settings are provided below:
\begin{itemize}
    %\item $\mathbf{T}$ \textbf{(same node)}: This is the model without the kinematic layer, which only uses trajectory information. All the nodes share the same embedding function.
    \item $\mathbf{T}$: This is the model without the kinematic layer, which only uses trajectory information.
    %\item $\mathbf{T}$: This is the model which uses WAE but without the kinematic constraint layer and only has trajectory information.
    \item $\mathbf{T}+\mathbf{C}-\mathbf{ATT}$: This is the model without the dual-attention mechanism or the kinematic constraint layer. We used equal attention in this model setting instead.
    \item $\mathbf{T}+\mathbf{C}$: This is the model without kinematic constraint layer.
    \item $\mathbf{T}+\mathbf{C}+\mathbf{K}$: This is the whole proposed model including all the components.
\end{itemize}
The ADE / FDE of each model setting are shown in the lower part of Table \ref{tab:ID1}.

$\bullet$ $\mathbf{T}$ versus $\mathbf{T}+\mathbf{C}$:
We show the effectiveness of employing scene context information.
$\mathbf{T}$ is the model without the kinematic layer, which only uses trajectory information, while $\mathbf{T}+\mathbf{C}$ further employs context information. 
The models directly output the position displacements $(\Delta x^k, \Delta y^k)$ at each step, which are aggregated to get complete trajectories. 
We can see little difference on prediction errors over short horizons, while the gap becomes larger as the horizon extends. The reason is that the vehicle trajectories within a short period can usually be well approximated by a constant velocity model, which are not heavily restricted or affected by the static environmental context. 
However, as the forecasting horizon increases, the effects of context constraints cannot be ignored anymore, which leads to larger performance gain of leveraging context information.
Compared with $\mathbf{T}$, the 5.0s ADE / FDE of $\mathbf{T}+\mathbf{C}$ are reduced by 5.6\% / 6.4\%, 9.3\% / 7.7\% and 2.1\% / 2.4\% in RA, UI and HR scenarios, respectively. This implies that the context information has larger effects on the prediction in RA and UI scenarios, where the influence of road geometries cannot be ignored.
The context information does little help to HR scenarios, since most vehicles go straight on highways.
In Fig. \ref{fig:ID_plot}, the predicted distribution of $\mathbf{T}+\mathbf{C}$ is more compliant to roadways to avoid collisions and the vehicles near the ``yield'' or ``stop'' signs tend to yield or stop.
However, $\mathbf{T}$ generates samples which are outside of feasible areas or violating traffic rules.

$\bullet$ $\mathbf{T}+\mathbf{C}-\mathbf{ATT}$ versus $\mathbf{T}+\mathbf{C}$:
We show the effectiveness of the proposed dual-attention mechanism.
$\mathbf{T}+\mathbf{C}-\mathbf{ATT}$ uses equal attention coefficients in both topological and temporal layers. 
According to the statistics reported in Table \ref{tab:ID1}, compared with equal attention, employing the dual-attention mechanism to figure out relative importance within the topological structure and along different time steps can reduce the 5.0s ADE / FDE by 21.3\% / 25.0\%, 27.9\% / 21.5\% and 7.8\% / 8.0\% in RA, UI and HR scenarios, respectively.
The improvement in RA and UI are more significant due to frequent interactions.

$\bullet$ $\mathbf{T}+\mathbf{C}$ versus $\mathbf{T}+\mathbf{C}+\mathbf{K}$: 
We show the effectiveness of the kinematic constraint layer.
Different from $\mathbf{T}$ which directly output position displacement, the outputs of the GRU unit in $\mathbf{T}+\mathbf{C}+\mathbf{K}$ are control actions, which are aggregated by the bicycle model to obtain complete trajectories.
According to Table III, employing the kinematic constraint layer to regularize the learning-based prediction hypotheses can further reduce the 5.0s ADE / FDE by 20.0\% / 13.7\%, 12.5\% / 16.0\% and 9.5\% / 9.3\% in RA, UI and HR scenarios, respectively.
Due to the restriction from the kinematic model, unfeasible movements can be filtered out and the model is unlikely to overfit noisy data or outliers.
Moreover, the improvement in RA and UI is more significant than in HR. The reason is that most vehicles go straight along the road in HR, whose behaviors can be well approximated by linear models. However, there are frequent turning behaviors in RA and UI which need constraints by more sophisticated models. 
We also visualize the predicted trajectories in Fig. \ref{fig:ID_plot}, where the ones in $\mathbf{T}+\mathbf{C}+\mathbf{K}$ are smoother and more plausible.

\section{Conclusion}
In this paper, we propose a generic system for multi-agent trajectory prediction named STG-DAT, which considers context information, trajectories of heterogeneous, interactive agents and physical feasibility constraints. In order to effectively model the interactions between different entities, we design a graph dual-attention network to extract features from spatio-temporal dynamic graphs. The Wasserstein generative modeling is employed as the basis of training the whole framework. The STG-DAT is validated by both pedestrian and vehicle trajectory prediction tasks on multiple benchmark datasets. The experimental results show that our approach achieves the state-of-the-art prediction performance compared with multiple baseline methods. Moreover, the proposed prediction model can be easily adopted by multi-target tracking frameworks, which empirically proves to enhance tracking accuracy.

% or


\appendices
\section{Data Preprocessing}
\begin{figure}[!tbp]
	\begin{center}
		\includegraphics[width=\columnwidth]{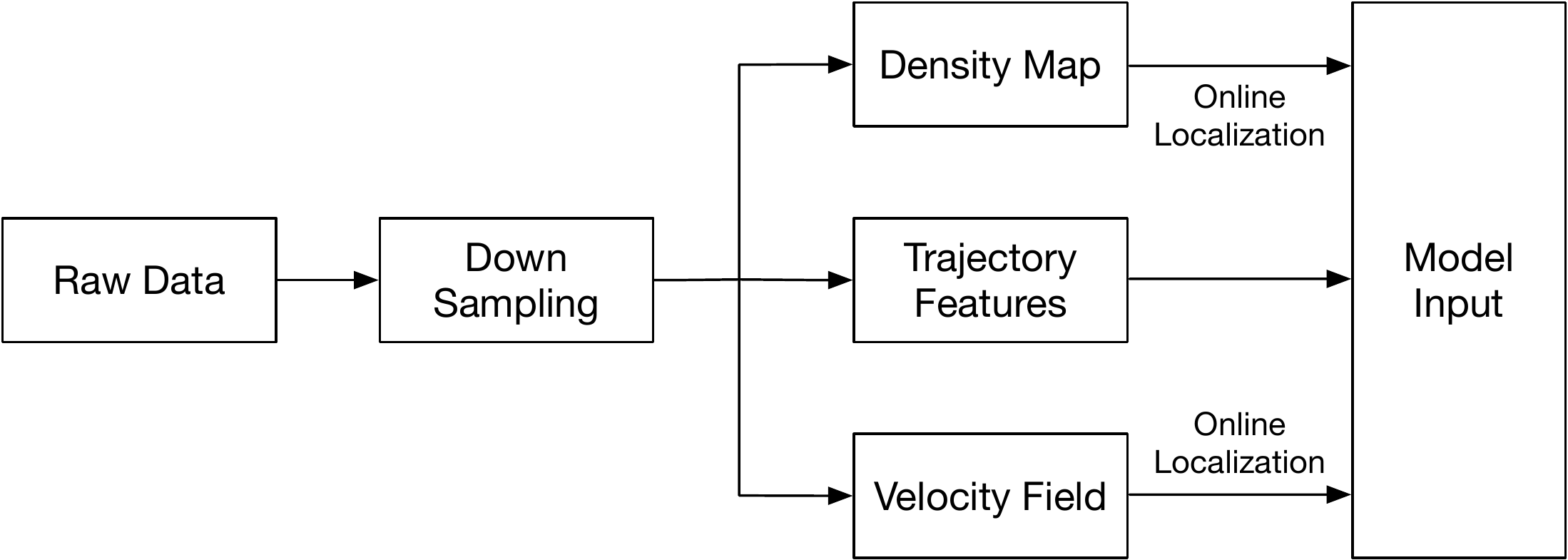}
		\caption{The pipeline of data preprocessing. The raw data was first down-sampled to be compatible to our experiment setup.}
		\label{fig:preprocessing}
	\end{center}
\end{figure}

In this section, we introduce the supplementary details of data preprocessing. The pipeline is illustrated in Fig. \ref{fig:preprocessing}.

\subsection{Global Context Information}
In order to provide better global context information, we designed two different representations, namely occupancy density map and mean velocity field. After constructing such global context information offline, we did decentralized online localization for the corresponding target agent and obtained their local context information, which was used in both training and testing phases. 
\vspace{0.1cm}

\noindent\textbf{Occupancy Density Map}
The density map describes the normalized frequency distribution of all the agents' locations. For a specific scene, we first split our map into a number of bin areas, which are 1m$\times$1m squares. Without loss of generality, we denote this histogram as $B$, and all the agents in different frames as a set $\{o_{k,p}\}$, where $p$ is the agent index and $k$ is the frame index. 
We obtained the global representation of density by calculating $B_{i,j}=\sum_{t,k}\phi(o_{k,p},i,j)$, where $i,j$ are the indices of the histogram and $\phi(o_{k,p},i,j)$ is an indicator function which equals 1 if $o_{k,p}$ is located in the bin area indicated by index $i,j$ and 0 otherwise. Then we normalized this density map by dividing all bin values by the sum of the values in this histogram and used this normalized histogram as our occupancy density map.
\vspace{0.1cm}

\noindent\textbf{Mean Velocity Field}
Similarly, we also created a map of velocity field which contains 1m$\times$1m square areas. We denote the whole map as $VF$ and the bin item indexed by $i,j$ as $VF(i,j)$. The $VF(i,j)$ is a two-dimensional vector representing the average speed along vertical and horizontal axes of all the agents in this area. 
More formally, 
\begin{equation}
	\begin{aligned}
	VF(i,j)_{x} = \frac{1}{N}\sum_{k,p}\phi(v_{k,p},i,j)v_{k,p}^{x}, \\ 
	VF(i,j)_{y} = \frac{1}{N}\sum_{k,p}\phi(v_{k,p},i,j)v_{k,p}^{y},
	\end{aligned}
\end{equation}
where $N$ is the number of points located at the bin area $(i,j)$.

\subsection{Localization for Local Context Information}
After obtaining the global context offline, our model utilized a decentralized method to do localization for each agent during training and testing. Given the location and the moving direction of the current agent at the current time step, we obtained a local context centered on this agent along its moving direction from the global context. All the agents share the same size of the local context map.
%Figure \ref{fig:local-central} provides an illustrative example.

%\begin{figure}[htbp]
%\begin{center}
%\includegraphics[width=\columnwidth]{central.pdf}
%\caption{The illustrative diagram of local context information. The target agent is denoted by the red star and its local context is the $3\times 3$ matrix denoted by the red box centered on itself.}
%\label{fig:local-central}
%\end{center}
%\end{figure}

% % you can choose not to have a title for an appendix
% % if you want by leaving the argument blank
% \section{}

% % use section* for acknowledgment
% \section*{Acknowledgment}


\ifCLASSOPTIONcaptionsoff
  \newpage
\fi

\bibliographystyle{IEEEtran}
%% argument is your BibTeX string definitions and bibliography database(s)
\bibliography{reference}

\end{document}